%% file: main.tex
\documentclass[10pt,twocolumn,letterpaper]{article}

\usepackage[pagenumbers]{cvpr} 
\usepackage{bm}
\usepackage{float}
\usepackage{xcolor}


\definecolor{cvprblue}{rgb}{0.21,0.49,0.74}
\usepackage[pagebackref,breaklinks,colorlinks,citecolor=cvprblue]{hyperref}

\def\paperID{6} 
\def\confName{CVPR}
\def\confYear{2024}

\newcommand{\MEp}{me_p }
\newcommand{\MEv}{me_v }
\newcommand{\temporalWeight}{\lambda_{vel} }
\newcommand{\orientation}{\theta_{g} }
\newcommand{\bodyPose}{\theta_{p} }
\newcommand{\bodyPoseOrientWeight}{\beta }
\newcommand{\translation}{\kappa}
\newcommand{\scale}{\sigma}
\newcommand{\birdModel}{M}

\newcommand{\matr}[1]{\mathbf{#1}}
\newcommand{\vecn}[1]{\mathbf{#1}}
\newcommand{\gvecn}[1]{\boldsymbol{#1}} 

\title{Temporally-consistent 3D Reconstruction of Seabirds}

\author{Johannes Hägerlind$^1$, Jonas Hentati-Sundberg$^2$, Bastian Wandt$^1$\\
$^1$Linköping University, Sweden\\
$^2$Swedish University of Agricultural Sciences, Sweden\\
{\tt\small \{johannes.hagerlind, bastian.wandt\}@liu.se, jonas.sundberg@slu.se}
}

\begin{document}
\maketitle

\begin{abstract}
This paper deals with 3D reconstruction of seabirds which recently came into focus of environmental scientists as valuable bio-indicators for environmental change.
Such 3D information is beneficial for analyzing the bird's behavior and physiological shape, for example by tracking motion, shape, and appearance changes. 
From a computer vision perspective
birds are especially challenging due to their rapid and oftentimes non-rigid motions.
We propose an approach to reconstruct the 3D pose and shape from monocular videos of a specific breed of seabird -- the common murre.
Our approach comprises a full pipeline of detection, tracking, segmentation, and temporally consistent 3D reconstruction.
Additionally, we propose a temporal loss that extends current single-image 3D bird pose estimators to the temporal domain.
Moreover, we provide a real-world dataset of 10000 frames of video observations on average capture nine birds simultaneously, comprising a large variety of motions and interactions, including a smaller test set with bird-specific keypoint labels.
Using our temporal optimization, we achieve state-of-the-art performance for the challenging sequences in our dataset\footnote{\url{https://huggingface.co/datasets/seabirds/common_murre_temporal}}.
\end{abstract}

\section{Introduction and Related Work}
\label{sec:intro}
Studying detailed behaviour of animals is a fundamental topic in biological, ecological and environmental conservation research \cite{couzin2023emerging}. 
Seabirds are a large and diverse group of animals with a high conservation value and known for their potential to indicate changes in marine and terrestrial ecosystems \cite{elliott2008seabird, monaghan1996relevance}. Behavioural studies of seabirds has a long history, where novel technologies such as cameras and computer vision has been increasingly used in applied research \cite{edney2021applications, hentati2023seabird}.

An automated 3D reconstruction of searbirds from video sequences can offer detailed insights into behavior, physiology, and adaptability over time.
In this paper, we present a novel approach aimed at reconstructing the 3D pose and shape of a specific breed of seabird, namely the common murre (uria aalge). 
Our method encompasses a multi-stage pipeline, including detection, tracking, segmentation, and temporally consistent 3D reconstructions. 

\begin{figure}
    \centering
    \includegraphics[width=0.98\linewidth]{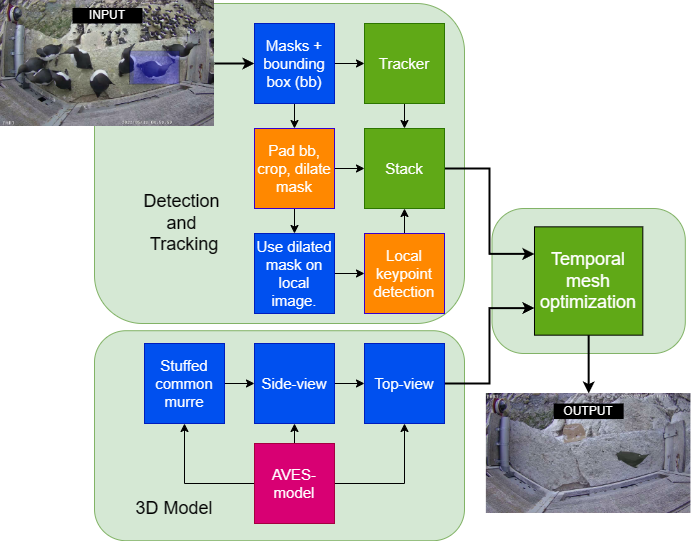}
    \caption{The proposed pipeline. The pink box represents learning the 3D pose prior \cite{wang2021birds}, the blue boxes introduce the fitting the parameterized model to the 3D fitting and the prediction of segmentation masks inspired \cite{hagerlind20233d}, the orange boxes additional improvements that were made in the current work, and the green boxes show the integration of temporal information which is the main contribution of this work.}
    \label{fig:pipeline}
\end{figure}

Many methods investigate the use of parametric mesh models to do 3D reconstruction of humans, \eg. methods that build upon SMPL \cite{smpl}, such as \cite{smplify,kanazawa2018end,zhang2020learning,kolotouros2021probabilistic,baradel2021leveraging,yuan2022glamr,tian2023recovering,sun2023trace,yao2024staf}). 
For birds \citet{badger20203d} develop a 3D reconstruction for cowbirds. 
\citet{wang2021birds} build on \cite{badger20203d} and developed species-specific as well as multi-species shape models. 
\citet{hagerlind20233d} noted that the method of \cite{wang2021birds} was not sufficient to reconstruct the \textit{common murre} from top-view images, which is the dominant view for the cliff-inhabiting common murre.
They use the pose prior and bone length prior of the cowbird model in \cite{badger20203d} to fit keypoints annotated in a 3D scan. The resulting bone length and shape parameters are used as an initialization for a more information-rich side-view optimization that uses 2D images annotated with keypoints and masks as input. 
In the side-view optimization \cite{hagerlind20233d} uses a similar method as in \cite{wang2021birds} and moved the mean of the bone length and the shape parameters towards that of the common murre. 
Finally, the results from the the side-view optimization were used to initialize the top-view optimization.

We build on top of the work by \cite{hagerlind20233d} by using the mesh parameters and optimization parameters and extending the
single image-based approach to a temporal approach. 
To achieve this we introduce a motion consistency assumption.
This temporal assumption is crucial for capturing the dynamic nature of seabird movements and ensuring the fidelity of reconstructed 3D poses over time. We also investigate the use of temporally consistent bone lengths. 
Additionally, to improve the keypoint detections, we investigate the use of a weighted median filter.
Fig.~\ref{fig:pipeline} shows our full framework.

To facilitate further research and benchmarking efforts in this domain, we introduce a real-world dataset comprising video observations with 10K consecutive frames, created by researchers in the Baltic Searbird Project \cite{baltic_seabird_project}. 
This dataset captures, on average, nine seabirds simultaneously engaged in a diverse array of behaviors, which lead to large pose changes, e.g. flapping their wings, and interactions with strong occlusions. We provide this dataset and a small test dataset containing keypoint labels for 7 birds in 100 consecutive frames at \url{https://huggingface.co/datasets/seabirds/common_murre_temporal}.

In summary, this paper presents a comprehensive framework for 3D reconstruction of seabirds from monocular videos, addressing the unique challenges posed by their behavior and movements. 
Through our proposed method and the accompanying dataset, we aim to advance the field of seabird research, providing valuable insights into their ecological significance and responses to environmental change.

\section{Method}
\label{sec:method}
Fig.~\ref{fig:pipeline} shows all processing steps of our full approach.
It consists of a detection and tracking stage, an offline 3D scan fitting and the temporal pose optimization.

\subsection{Detection and Segmentation}
\label{sec:detection}
We use the segmentation network provided by \citet{alvarez2021alternative}. 
The keypoint detector is trained using DeepLabCut \cite{Mathisetal2018_deeplabcut, NathMathisetal2019} by fine-tuning a Resnet50 \cite{resnet_he2016deep}. 
The training dataset consists of 500 images with 20 keypoints (2 more than \cite{hagerlind20233d}). 
Since there are many frames where birds are close together we follow \cite{hagerlind20233d} and consider each animal individually. 
First, the image is cropped using the bounding boxes obtained from the segmentation network. 
This is followed by masking all pixels that are not labeled by the predicted segmentation masks. 
To compensate for possible inaccurate segmentation masks, we pad the bounding box by 40 pixels in each direction and then dilate the original prediction using a squared kernel of width 70 as in \cite{hagerlind20233d}.

\subsubsection{Weighted Median Filter}
To filter occasional misdetections, a weighted median filter is applied to the detected 2D keypoints using a window size of 5.
The x and y coordinates are filtered separately. 
The coordinates are chosen based on the median of the cumulative sum of the confidence associated with the keypoints (separately for the x and the y dimensions). 
This reduces the amount of outliers in the keypoint detection.

\subsection{Tracking}
The tight bounding boxes around the predicted segmentation mask are used as input to a tracker. Using the bounding box of the segmentation masks allows for a direct connection between the tracker and the segmentation mask (necessary for later steps). In case a segmentation mask is missed, there is a 5-frame memory that keeps track of the previous prediction. 
We track based on the highest IoU between bounding boxes in consecutive frames.

\subsection{Fitting the 3D Model to the Image}
We aim to fit a 3D bird model to the 2D keypoint and 2D masks. 
To allow for batch-optimization we pad and scale the keypoints and segmentation masks to a dimension of 256x256 pixels. 
The starting point is the common murre model from \citet{hagerlind20233d} adapted from \cite{wang2021birds}. 
The shape and pose of the reconstructed bird model is controlled by the translation ($\translation$), the scale ($\scale$), the global orientation ($\orientation$), and the body pose $\bodyPose$ parameterized by joint angles. 
The scale parameter scales all the bones by a common factor. 
As in \cite{hagerlind20233d} we keep the depth fixed since the camera is looking from the top towards a flat surface. 
We keep the bone length constant since this was shown to reduce the perceptual quality in this setting (cf.~\cite{hagerlind20233d}).
The model $\birdModel$ is hence described by the function $\birdModel(\translation, \scale, \orientation, \bodyPose)$
As initialization, we use the method in \cite{hagerlind20233d} where we rotate the 3D bird in top-view by 360° in 12° steps and select the one that best matches the predicted 2D keypoints from Sec.~\ref{sec:detection}.
We optimize the full parameter set $\translation, \scale, \orientation$, and $\bodyPose$. 

\noindent\textbf{Frame-wise objective.} 
We minimize the frame-wise loss from \cite{hagerlind20233d}
that achieved the best results in \cite{hagerlind20233d}:
\begin{equation}
E_{start}(\Theta) =
    \lambda_{kpt}  E_{kpt} +
    \lambda_{msk}  E_{msk} +
    \lambda_{pp}   E_{pp},
\end{equation}
where $E_{kpt}$ is a keypoint reprojection error, $E_{msk}$ is a mask error, and $E_{pp}$ is a pose prior.
We set $\lambda_{msk} =1$, $\lambda_{kpt} = 1$, and $\lambda_{pp} = 100$.
The keypoint loss, mask loss, and pose prior loss are calculated similar to \cite{wang2021birds}. 
The keypoint loss is an instance of the Geman-McLure error function (cf. \cite{geman1987statistical}) given by

\begin{equation}
    E_{kpt} = \sum_{i=1}^N c_i \frac{\sigma^2 (\Pi(m_{i})-p_{i})^2}{\sigma^2 + (\Pi(m_{i})-p_{i})^2},
\end{equation}
where $N$ is the number of keypoints, $\Pi(m_i)$ is a projected keypoint from the mesh (using a simple perspective camera without any distortion) and $p_i$ is the corresponding target keypoint. $c_i$ is the confidence assigned to a keypoint prediction. 
As in previous work \cite{wang2021birds, hagerlind20233d} we use $\sigma = 50$.
The mask loss $E_{mask}$ is calculated as the L1 distance between the predicted mask and the soft mask (silhouette) using PyTorch soft rasterizer \cite{pytorch3d_ravi2020}.
The pose prior loss is calculated using the squared Mahalanobis distance as in \cite{bishop2006pattern}:

\begin{equation}
    E_{pp} = (\vecn{x}-\gvecn{\mu})^T \matr{\Sigma^{-1}} (\vecn{x}-\gvecn{\mu}),
\end{equation}
where the mean $\gvecn{\mu}$ is taken from \cite{hagerlind20233d}, and the the covariance $\matr{\Sigma}$ is taken from \cite{badger20203d} (from the cowbird species).

\noindent\textbf{Temporal objective.} 
Since our goal is to achieve temporal consistency in a sequence of poses, we introduce two additional regularization terms for the velocity and the acceleration. 

The first regularizer aims to decrease the difference between consecutive 3D poses 
\begin{equation}
E_{vel} = \sum_{k \in \{g, p\}} \bodyPoseOrientWeight_k \sum_{i=1}^N{\lVert \theta_{k, i+1} - \theta_{k, i} \rVert_2}
.
\end{equation}
While regularizing the velocity already significantly smoothes the motion, some jitters remain. 
To this end, we introduce another acceleration-based term:
\begin{equation}
E_{acc} = \sum_{k \in \{g, p\}} \bodyPoseOrientWeight_k \sum_{j=2}^N{\lVert \theta'_{k, j+1} - \theta'_{k,j} \rVert_2}
.
\end{equation}
$\theta'$ denotes velocity. The global orientation $\orientation$ and body pose $\bodyPose$ have separate weights: $\bodyPoseOrientWeight_g = 10$ for global orientation and $\bodyPoseOrientWeight_p=1$ for body pose. This is based on the assumption that movements in the joints are likely to be faster than global orientation changes.

The combined objective function is 

\begin{equation}
    \label{eqn:combined_objective}
    E = E_{start} + \lambda_{vel} E_{vel} + \lambda_{acc} E_{acc}
    .
\end{equation}

\noindent\textbf{Common size constraint.} 
Although a bird can vary in shape, the bone length should remain constant during a reasonable time frame. 
In some experiments, we enforce this by optimizing a single scale for all bones during the full temporal window.

\noindent\textbf{Optimization.}
There are two steps in the mesh optimization, excluding initialization.
The first step uses the objective in Eq.~\ref{eqn:combined_objective} and the second step adds a mask loss. The first step uses 600 iterations and the second step uses 400 iterations. We use the Adam optimizer \cite{Paszke_PyTorch_An_Imperative_2019} and a learning rate of 0.01.

\section{Experiments}
\label{sec:experiments}
\subsection{Dataset}
The common murre is a particularly interesting seabird as an indicator of environmental change since it heavily interacts with the environment by catching fish in the ocean.
Moreover, it is relatively easy to observe since it breeds on cliffs that can be equipped with surveillance cameras. Researchers in the Baltic Searbird Project \cite{baltic_seabird_project} have created a dataset comprising 10K consecutive frames capturing common murres on a cliff ledge during main breeding season. The resolution is $2592 \times 1520$px and the frame rate is 25 frames per second.
On average there are nine birds in the camera view.
We identify several different behaviors: standing, walking, flying away, approaching, preening, flapping wings, and attacking other birds.
It shows many challenging poses from bending the neck backward as well as non-rigid deformations, mainly of the neck.
Additionally, interactions between individual birds lead to strong occlusions posing an additional challenge for tracking and reconstruction.
In addition to the video sequences, we provide temporally consistent 2D keypoint labels for 100 images for 7 out of 9 birds for testing purposes.
While we target accurate and time-consistent 3D reconstruction, this dataset also enables further behavioral studies for the computer vision community.

\subsection{Metrics}
Since there is no available 3D data for evaluation, we use the 2D reprojection of the keypoints in the mesh and compare them with the ground truth evaluation. 
The $\MEp$, $\MEv$ measure the RMS error for the projected mesh keypoints position and velocity respectively. This is calculated by dividing the error by the longest side of the bounding box of the predicted segmentation mask to enable comparison in different scales.

\subsection{Experiments}

\begin{figure}
    \centering
    \includegraphics[width=0.98\linewidth]{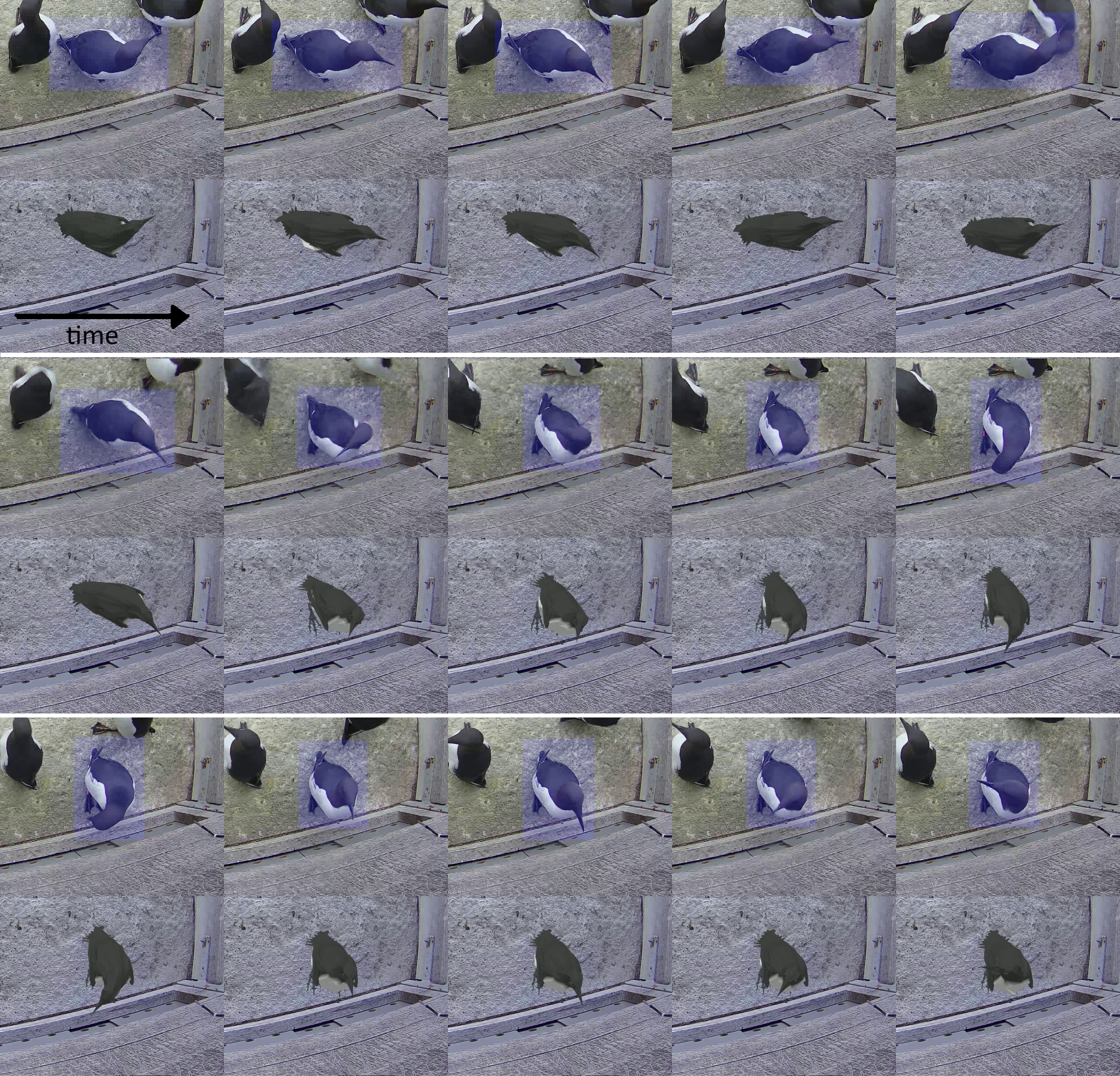}
    \caption{Example reconstruction. The odd rows show the input image. The even rows show the corresponding mesh for the tracked bird rendered on top of the background image. The texture of the reconstructed bird is only added for visualization purposes.
    }
    \label{fig:reconstruction}
\end{figure}

We conduct a line of experiments in different settings, evaluating all part of our proposed pipeline.
Fig.~\ref{fig:reconstruction} shows an example of a 3D reconstruction from our approach. 
Note that we only show reconstructions for a subset of all the birds in the image.
This is due to limitations in the segmentation network that only provides trackable regions for the visualized birds.
A single image reconstruction for the remaining frames is conceivable but here we focus on the results of our tracker in combination with our temporal optimization.
The supplementary material contains additional videos showing reconstructions on the test set using different parameter settings.

\subsection{Quantitative Results}
In total, 66 experiments were conducted. A temporal window of 1 (no temporal optimization) and 100 is investigated. For the window size of 1, the use of a median filter for the input 2D joints is investigated. 
The setting with window size 1 and no median filter corresponds to the setting used by \cite{hagerlind20233d}.

For the temporal window size of 100, the following cases are investigated:
\begin{itemize}
    \item $\lambda_{vel}, \in \{10^2, 10^3, 10^4, 10^5\}$
    \item Use acceleration loss: true/false. 
    If true $\lambda_{acc} = \lambda_{vel}$. If false $\lambda_{acc} = 0$
    \item Use weighted median filter (for the predicted keypoints) (True/False)
    \item Optimize a common size in the temporal window (True/False).
\end{itemize}

\begin{table}
\footnotesize
    \centering
    \begin{tabular}{lllll|cc}
     & $\temporalWeight$ & acc & med & size & $\MEp \downarrow$ & $\MEv \downarrow$ \\
    \hline
    baseline & 0 & - & False & - & 0.0824 & 0.0322 \\
    \hline
    Ours & $10^5$ & False & True & True & 0.1647 & 0.0164 \\
    & $10^4$ & False & True & False & 0.1099 & 0.0142 \\
    & 1000 & True & False & True & 0.0872 & 0.0196 \\
    & 1000 & True & False & False & 0.0845 & 0.0174 \\
    & 1000 & True & True & True & 0.0824 & 0.0172 \\
    & 1000 & False & False & False & 0.0823 & 0.0188 \\
    & 1000 & False & True & True & 0.0823 & 0.0179 \\
    & 1000 & False & False & True & 0.0820 & 0.0187 \\
    & 1000 & True & True & False & 0.0817 & 0.0154 \\
    & 1000 & False & True & False & 0.0816 & 0.0178 \\
    & 0 & - & True & - & 0.0809 & 0.0262 \\
    & 100 & False & True & False & 0.0804 & 0.0182 \\
    & 100 & False & True & True & 0.0804 & 0.0185 \\
    & 100 & False & False & False & 0.0803 & 0.0223 \\
    & 100 & False & False & True & 0.0803 & 0.0228 \\
    & 100 & True & False & False & 0.0793 & 0.0196 \\
    & 100 & True & False & True & 0.0791 & 0.0196 \\
    & \textit{100} & \textit{True} & \textit{True} & \textit{False} & \textit{0.0758} & \textit{0.0149} \\
    Ours (best) & \textbf{100} & \textbf{True} & \textbf{True} & \textbf{True} & \textbf{0.0756} & \textbf{0.0150} \\

    \end{tabular}
    \caption{Evaluation on our test set sorted in descending order (worst to best). The first row shows the baseline. The following abbreviations are used: \textit{acc} for acceleration loss, \textit{med} for the median filter, $\MEp$ for mean error of the keypoint positions, and $\MEv$ for mean error of the keypoint position velocities. Each individual contribution improves the performance.}
    \label{tab:evaluation}
\end{table}

Table~\ref{tab:evaluation} shows the evaluation results. 
The best $\MEp$ is achieved using a window size of 100, a temporal loss of 100, a common size in the window, and an acceleration loss. 
Each individual component improves the performance.
Looking at the top-8 $\MEp$ we see that using the weighted median filter in conjunction with the acceleration loss results in a lower $\MEp$. 
The best $\MEp$ for a window size of 100 is $6.6\%$ lower than the best result for a window size of 1, validating the superior performance of our temporal approach compared to single frame methods. 

Comparing $\temporalWeight=100$ and $\temporalWeight=1000$ we see that the former results in a better $\MEp$. The two last rows in the table show the best result for $\temporalWeight = 10^4$ and $\temporalWeight = 10^5$ respectively. Using either $\temporalWeight = 10^4$ and $\temporalWeight = 10^5$ greatly increases the $\MEp$ (\ie worsens the result).

Using $\temporalWeight = 0$, \ie window-size 1, produces many non-existing high-frequency motions for the body pose and global orientation. Furthermore, the scale of the bird changes in an unnatural way.

Using $\temporalWeight = 100$ reduces many of the non-existing high-frequency motions, but not all, and $\temporalWeight = 1000$ further reduces these motions.

A large weight for the temporal regularizer of $\temporalWeight \in \{10^4, 10^5\}$ fails to capture the quick motion of the birds and $\temporalWeight = 10^5$ even has a severe negative impact on the size even if optimizing a single size in the window (see videos in supplementary material).

\section{Conclusion}
This pilot study investigates how different temporal assumptions can be used to improve the 3D reconstruction of the common murre captured by monocular cameras. 
We showed that our temporal regularizer, including the acceleration, leads to a significantly improved performance when used together with our weighted median filter, which improves the 2D keypoint prediction. 
Additionally, the temporal loss helps to enforce more physically plausible motions. 
Moreover, optimizing for a single scale during the whole sequence is another way to enforce temporal coherence and further improves the reconstruction. 
Since we build upon \cite{wang2021birds} our method still fails for extreme pose changes. 

We will deal with such strong deformations in future work.

\section{Acknowledgments}

This work was partially supported by the Wallenberg AI, Autonomous Systems and Software Program (WASP) funded by the Knut and Alice Wallenberg Foundation, Sweden.

The computations were enabled by the Berzelius resource provided by the Knut and Alice Wallenberg Foundation at the National Supercomputer Centre.

{
    \small
    \bibliographystyle{ieeenat_fullname}
    \bibliography{main}
}

\end{document}